\documentclass[sigconf]{acmart}

\usepackage{graphicx}
\usepackage{hyperref}
\usepackage{url}
\usepackage{algorithm}
\usepackage{algorithmic}
\usepackage{multirow,makecell,bbm}
\usepackage{dsfont,bm}
\usepackage{lineno,listings}
\usepackage{xcolor}
\usepackage{amsmath}
\usepackage{balance}

\usepackage{xcolor}

\definecolor{codegreen}{rgb}{0,0.6,0}
\definecolor{codegray}{rgb}{0.5,0.5,0.5}
\definecolor{codepurple}{rgb}{0.58,0,0.82}
\definecolor{backcolour}{rgb}{0.95,0.95,0.92}

\lstdefinestyle{mystyle}{
    backgroundcolor=\color{backcolour},   
    commentstyle=\color{codegreen},
    keywordstyle=\color{magenta},
    numberstyle=\tiny\color{codegray},
    stringstyle=\color{codepurple},
    basicstyle=\ttfamily\footnotesize,
    breakatwhitespace=false,         
    breaklines=true,                 
    captionpos=b,                    
    keepspaces=true,                 
    numbers=left,                    
    numbersep=5pt,                  
    showspaces=false,                
    showstringspaces=false,
    showtabs=false,                  
    tabsize=2
}

\AtBeginDocument{%
  }

\setcopyright{acmlicensed}
\copyrightyear{2023}
\acmYear{2023}
\acmConference[MM '23]{Proceedings of the 31st
ACM International Conference on Multimedia}{October 29-November 3,
2023}{Ottawa, ON, Canada}
\acmBooktitle{Proceedings of the 31st ACM International Conference on
Multimedia (MM '23), October 29-November 3, 2023, Ottawa, ON, Canada}
\acmPrice{15.00}
\acmDOI{10.1145/3581783.3612059}
\acmISBN{979-8-4007-0108-5/23/10}

\begin{document}

\title{PBFormer: Capturing Complex Scene Text Shape with Polynomial Band Transformer}

% \affiliation{%

\author{Ruijin Liu}
\affiliation{%
  \institution{Xi'an Jiaotong University}
  \state{Xi'an}
  \country{China}
  }
\email{lrj466097290@stu.xjtu.edu.cn}

\author{Ning Lu}
\affiliation{%
  \institution{Huawei Technologies Ltd.}
  \state{Shenzhen}
  \country{China}
  }
% \email{luning12@huawei.com}

\author{Dapeng Chen}
\affiliation{%
  \institution{Huawei Technologies Ltd.}
  \state{Shenzhen}
  \country{China}
  }
% \email{chendapeng8@huawei.com}

\author{Cheng Li}
\affiliation{%
  \institution{Huawei Technologies Ltd.}
  \state{Shenzhen}
  \country{China}
  }
% \email{licheng81@huawei.com}

\author{Zejian Yuan}
\affiliation{%
  \institution{Xi'an Jiaotong University}
  \state{Xi'an}
  \country{China}
  }
% \email{yuan.ze.jian@xjtu.edu.cn}

\author{Wei Peng}
\affiliation{%
  \institution{Huawei Technologies Ltd.}
  \state{Shenzhen}
  \country{China}
  }

\renewcommand{\shortauthors}{Ruijin Liu et al.}
%% No italics and no comma
%% If needed use a foot or author note to identify equal contribution

\begin{abstract}
  We present PBFormer, an efficient yet powerful scene text detector that unifies the transformer with a novel text shape representation \textbf{P}olynomial  \textbf{B}and (PB).  The representation has four polynomial curves to fit a text's top, bottom, left, and right sides, which can capture a text with a complex shape by varying polynomial coefficients.  PB has appealing features compared with conventional representations: 1)  It can model different curvatures with a fixed number of parameters, while polygon-points-based methods need to utilize a different number of points.  2) It can distinguish adjacent or overlapping texts as they have apparent different curve coefficients, while segmentation-based or points-based methods suffer from adhesive spatial positions. PBFormer combines the PB with the transformer, which can directly generate smooth text contours sampled from predicted curves without interpolation. A parameter-free cross-scale pixel attention (CPA) module is employed to highlight the feature map of a suitable scale while suppressing the other feature maps. The simple operation can help detect small-scale texts and is compatible with the one-stage DETR framework, where no postprocessing exists for NMS. Furthermore, PBFormer is trained with a shape-contained loss, which not only enforces the piecewise alignment between the ground truth and the predicted curves but also makes curves' position and shapes consistent with each other.  Without bells and whistles about text pre-training, our method is superior to the previous state-of-the-art text detectors on the arbitrary-shaped text datasets.
\end{abstract}









% \renewcommand\footnotetextcopyrightpermission[1]{}
% \settopmatter{printacmref=false} %remove ACM reference format

\maketitle

\section{Introduction}

Scene text detection is an active research topic in computer vision and enables many downstream applications such as image/video understanding, visual search, and autonomous driving~\citep{TextSurvey,RoadText}. However, the task is also challenging. One non-negligible reason is that the text instance can have a complex shape due to the non-uniformity of the text font, skewing from the photograph, and specific art design. Capturing complex text shapes needs to develop effective text representation. State-of-the-art methods roughly tackle this problem with two types of representations. One is the point-based representation, which predicts the points on the image space to control the shape of the points, including the Bezier control points~\citep{ABCNet} and polygon points~\citep{TextBPN}. The other produces segmentation maps. The map can describe the text of various shapes and can benefit from the prediction results at the pixel level~\citep{DBNet,FCENet}. 

Despite the good performance, both types of representation have limitations: 1) Points-based methods suffer from a fixed number of control points~\citep{FewCouldBeBetter,TESTR}. Too few points cannot handle the highly-curved texts, while simply adding points will increase redundancy for most perspective texts. 2) Segmentation-based methods needs post-processing and often requires extensive training data~\citep{FCENet}. 3) They both frequently fail in dividing adjacent texts due to ambiguous spatial positions of points or segment masks.

\begin{figure*}[t]
\includegraphics[width=177mm]{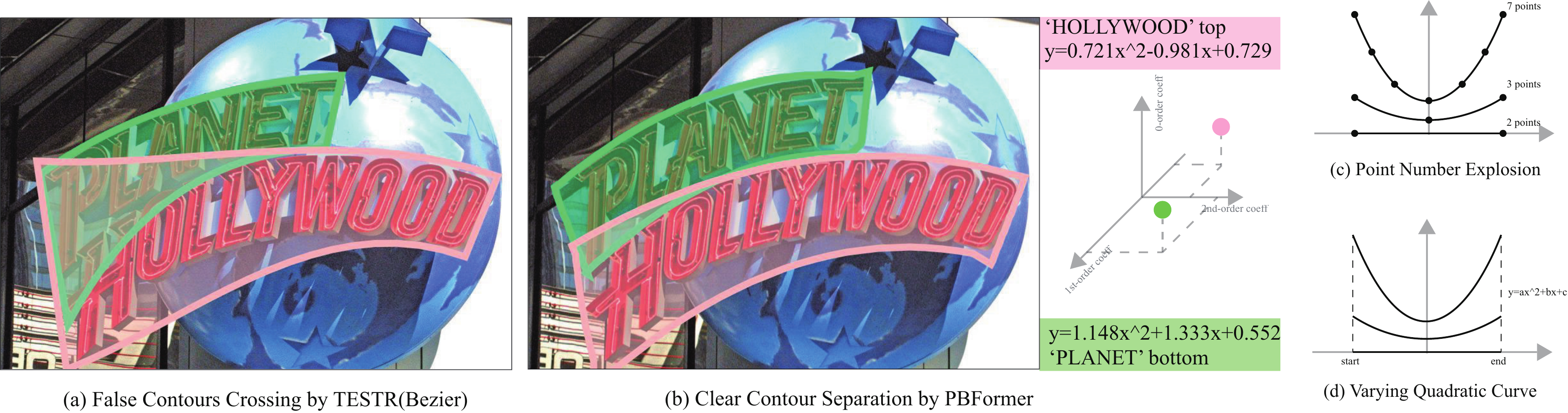}
\vspace{-1.1em}
\caption{\textbf{Advantages of PB over Bezier control points and polygon points.} Comparing (a) and (b), PB divides adjacent texts more clearly than Bezier control points. (c) shows the number of output point increase gradually to represent shapes from straight to highly curved. (d) shows varying curve coefficients can handle dynamic shapes with fixed number of variables. }
\label{fig:intro} 
\vspace{-1.1em}
\end{figure*}

To address these limitations, we propose a novel representation, named \textbf{P}olynomial \textbf{B}and (PB). It has clear advantages compared with previous text representations. In particular, PB consists of four polynomial curves, each of which fits along a text's top, bottom, left, and right sides. 
First, the coefficients of PB are discriminative in the parameter space even if the two texts are very close in the image space, as shown in Fig.~\ref{fig:intro}. 
Second, PB is represented by the functions defined in the image space. We can directly compare the ground truth contour points with the sampled points from polynomial curves by re-sampling techniques. 
It differs from Bezier-curve-based methods that need to generate the intermediate representation,  \emph{i.e.}, ``control points'' for supervision. 
The loss defined by the control points cannot truly reflect how humans percept the shape. A small difference in control points will lead to a large shape difference.

Witnessing the great success in NLP~\citep{AttentionIsAllYourNeed}, there has been a recent surge of interest in introducing transformers to vision tasks, including scene text detection. The current transformer-based text detectors are with two stages, such as FewBetter~\citep{FewCouldBeBetter} and  TESTR~\citep{TESTR}. We apply the proposed PB to the one-stage deformable DETR to improve the efficiency. In particular, we design parameter-free cross-scale pixel attention (CPA) module between the CNN feature and the transformer encoder-decoder layers. The CPA module first aligns the feature maps of different scales by enlarging all the feature maps to the same scale. Then it performs the cross-scale attention that highlights the feature value from a suitable scale while suppressing the other feature maps. With the scale-selective mechanism, our method becomes more compatible with the transformer decoders that do not have NMS for postprocessing. It implicitly suppresses the text proposals with incorrect scales, alleviating the learning burden of the transformer encoder-decoder layers. The features from CPA are effective to represent the shape of the text, two layers of transformer encoder-decoders are sufficient to detect the reasonable size of PB without NMS.

% For example, FewBetter~\citep{FewCouldBeBetter} adopts a CNN network to segmentation masks, then incorporates a transformer to generate the control points of the polygon or Bezier-Curve. TESTR~\citep{TESTR} also decodes the control point in each region of interest after extracting the bounding box proposals. They sacrifice efficiency due to the separated stages and destroy the simplicity of the detection transformer scheme.

The transformer decodes each polynomial curve's $K$ coefficients and $2$ boundary variables that determine the curve's definition domain. We uniformly sample the points on the predicted curves within the definition domain and compare them with the corresponding points on the ground truth polygon. Such a design supervises the curve piece by piece and can learn the curve shape and range consistently. In summary, the contribution of PBFormer is:

% We equip the transformer with PB. Instead of adopting the two-stage pipeline, we insert a parameter-free cross-scale pixel attention module between the CNN feature and the transformer encoder-decoder layers. The module enlarges CNN feature maps and generates multi-scale attention maps. After the attention module, we can directly send the per-pixel features to the transformer without inserting a region proposal module or grouping the segmentation results. The transformer decodes each polynomial curve's $K$ coefficients and $2$ boundary variables that determine the curve's definition domain. We uniformly sample the points on the predicted curves within the definition domain and compare them with the corresponding points on the ground truth polygon. Such a design supervises the curve piece by piece and can learn the curve shape and range consistently. In summary, the contribution of PBFormer is:
\begin{itemize}
  \item A novel text representation called the Polynomial Band (PB) is proposed. PB can utilize a fixed number of parameters to capture the text instance with various curvatures. It also excels at distinguishing the spatially close text instances. 
  \item  A cross-scale pixel attention module is proposed. The module performs pixel-wise attention across the feature maps with different sizes. It implicitly highlights the text regions and enables the transformer to direct take all pixel-wise features as input. 
  \item We design a shape-constrained loss function. The loss enforces the piece-wise supervision over the predicted curve and consistently optimizes the curve coefficients and definition domains. 
\end{itemize}

Experiments on multiple multi-oriented and curved text detection datasets demonstrate the effectiveness of our approach. 
Without any pre-training on large-scale text datasets, our method achieves the best F-measure. Moreover, due to the lightweight network architecture, our method runs real-time and 4.4 $\times$ faster than other open-sourced transformer text detectors.

\section{Related Work}

% \paragraph{Text representation.}
\noindent \textbf{Text representation.}
Conventional text representation can be roughly divided into point-based and segmentation-based methods. With the need to capture more complex text shapes, the point-based methods gradually involve more points for text representation, including 2 points bounding box~\citep{BoxRep}, 4 points quadrilateral~\citep{QuadRep}, 16 points polygon~\citep{DRRG} or 30 points polygon~\citep{DRRG} changed by texts' length.
ABCNet~\citep{ABCNet} calculates Bezier control points (8 points), which is sufficient for most quadrilateral or slightly-curved texts but still suffers from highly-curved. One deficiency of point-based methods is that they usually predict a fixed number of points, which is hard to balance the performance between simple perspective texts and text instances with complex shapes~\citep{TESTR}. 
The other is discrete point regression in images is easily disturbed by noise such as occlusion, making it difficult to distinguish adjacent text clearly.
ESIR~\citep{ESIR} adopts a single polynomial that fits the text center line. However, they can only model horizontal texts and still output additional points to represent text contours.
In contrast, the polynomial curves in polynomial band can handle various orientations and shapes, from a straight line to a round curve with fewer and a fixed number of parameters.  

Segmentation-based representation can naturally handle complex-shaped texts due to pixel-wise description~\citep{PSENet, MaskTextSpotter, PAN, FCENet, DBNet, DBNetpp,PixelEmbedding}. However,  they frequently fail to divide adjacent texts due to ambiguous spatial positions. Although CentripetalText~\citep{CentripetalText} tackles this problem by detecting a shrunk text mask and reconstructing the contour by shift map, it suffers from high computation complexity. The proposed polynomial band overcomes this problem by considering the global curve shape. Curve coefficients in parameter space can be easily separated even though two texts are close, making more clear bounds in the text crowding scenario. 

% \paragraph{Text detection transformer.}
\noindent \textbf{Text detection transformer.}
There is a trend to equip the transformer~\citep{AttentionIsAllYourNeed} with scene text detection. Current methods~\citep{FewCouldBeBetter, TESTR} directly combine DETR variants~\citep{DeformableDETR} with point representation such as 16-point polygons or Bezier control points, and adopt the two-stage architecture for the ease of optimization. For example, FewBetter~\citep{FewCouldBeBetter} first extracts segmentation maps by CNN-FPN to show representative text regions, then samples feature points in each region and feed them into a transformer to further decode control points. TESTR~\citep{TESTR} follows~\citep{BoundaryNet, ContourNet} to detect the bounding boxes, then utilize a transformer to find the control points inside box. They sacrifice efficiency due to the two-stage pipelines and destroy the simplicity of the detection transformer scheme. Our PBFormer inherits the single-stage simplicity and inserts a parameter-free cross-scale pixel attention module between the CNN feature and transformer encoder-decoder layers. With the attention module, per-pixel features are fed into the transformer without generating any proposals, making the whole architecture more efficient.

\section{Methodology}

\begin{figure*}[t]
\includegraphics[width=135mm]{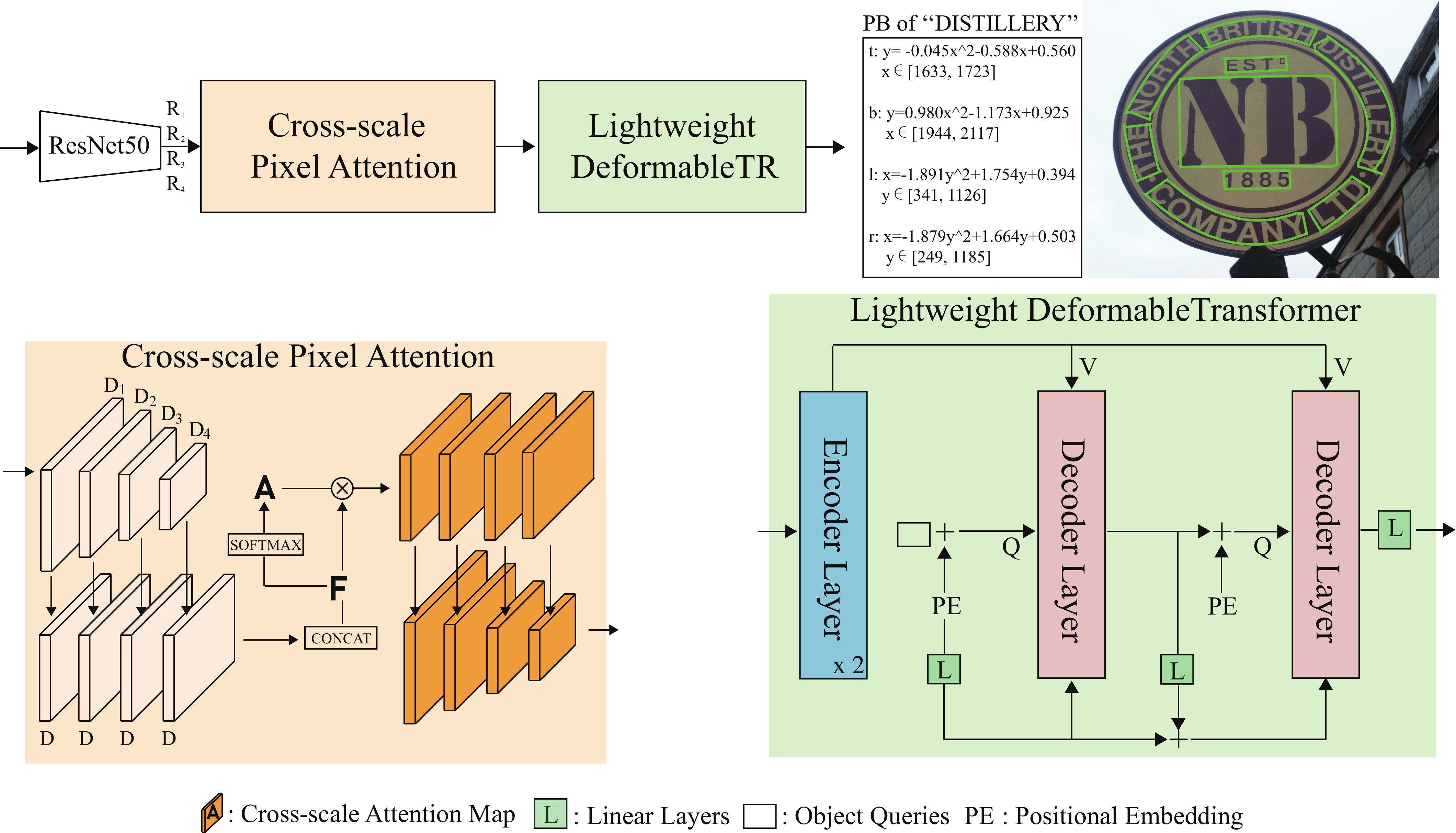} 
\vspace{-1.1em}
\caption{\textbf{Architecture of PBFormer.} The $\otimes$ and $+$ are element-wisely multiplication and addition, respectively. In the lightweight Deformable-Transformer, the object queries (white box) and Positional Embeddings (PE) are all learned parameters.}
\label{fig:network}
\vspace{-1.1em}
\end{figure*}

The overall framework of PBFormer is illustrated in Fig. \ref{fig:network}. Given an image with texts, PBFormer first employs a ResNet50 to produce the multi-scale feature maps, then feed the feature maps to a parameter-free cross-scale pixel attention module to highlight the texts' context information. The enhanced feature maps are concatenated and fed to a lightweight transformer, predicting the PBs' parameters. PB utilizes four polynomial curves to represent the shape of the text instance. It is simple but effective to capture different forms of texts. In the training phase, we sample dense points from the curves for PB parameters estimation according to both predicted curve coefficients and domain variables. A shape-constrained loss is designed to supervise curves piece by piece. 

\subsection{Polynomial Band}
% \label{PBdef}
We utilize four polynomial curves to represent the text instance's top, bottom, left, and right sides. The top and bottom boundaries are represented by $y=f^t\left(x\right)$  and $y=f^b\left(x\right)$:
\begin{equation}
\begin{split}
    y&=f^t\left(x\right)=a_2^tx^2 + a_1^tx + a_0^t,\quad x\in\left[e_0^t, e_1^t\right],\\
    y&=f^b\left(x\right)=a_2^bx^2 + a_1^bx + a_0^b,\quad 
x\in\left[e_0^b, e_1^b\right],
\end{split}
\end{equation}
where $a_2^t, a_1^t, a_0^t, a_2^b, a_1^b, a_0^b$ are polynomial coefficients. $(x, y)$ is the coordinate of a point on the boundary. $[e_0^t, e_1^t]$ and
$[e_0^b, e_1^b]$ are range of $x$ variable. 

One critical problem is that the polynomial curve is a \textit{single-value function} which means one point in the definition domain has a unique value in the value domain. The functions of the curves along the horizontal direction cannot be used to represent the curves along the vertical direction. For example, as Fig.~\ref{fig:network} shows, the text 'DISTILLERT's left (or right) side would not be represented by any $y=f\left(x\right)$. To solve it, we utilize  $x=f^l\left(y\right)$ and $x=f^r\left(y\right)$ to represent the left and right polynomial curves of the text instance: 
\begin{equation}
\begin{split}
\label{eq:leftcurve}
    x&=f^l\left(y\right)=a_2^ly^2 + a_1^ly + a_0^l,\quad y\in\left[e_0^l,e_1^l\right], \\
   x&=f^r\left(y\right)=a_2^ry^2 + a_1^ry + a_0^r,\quad y\in\left[e_0^r,e_1^r\right],
\end{split}
\end{equation}
where $[e_0^l, e_1^l]$ and $[e_0^r, e_1^r]$ define the range of $y$ variable.

\noindent \textbf{Output definition.}
We use $y=f^t\left(x\right),y=f^b\left(x\right),x=f^l\left(y\right),x=f^r\left(y\right)$ four polynomial curves that denote a band to wrap a text instance. Thus, the output is a 20-tuple $\theta$ that consists of all polynomial coefficients and boundary variables in the form of:
\begin{equation}
\label{PBdef}
    \theta = \left(a_2^t,a_1^t,a_0^t,e_0^t,e_1^t,a_2^b,a_1^b,a_0^b,e_0^b,e_1^b,a_2^l,a_1^l,a_0^l,e_0^l,e_1^l,a_2^r,a_1^r,a_0^r,e_0^r,e_1^r\right),
\end{equation}
where $a_2^t, a_2^b, a_2^l, a_2^r \neq 0$, $e_0^t, ... \in \left[0, 1\right]$, and all of them are real numbers. 

\subsection{Network Architecture.}
The network contains three modules: a ResNet-based CNN encoder, a parameter-free cross-scale pixel attention module, and a lightweight deformable transformer. We now introduce more details about the attention module and the lightweight transformer.

\noindent \textbf{Cross-scale pixel attention.}
The motivation of cross-scale pixel attention is to highlight text features at the best scale and suppress the others. Such a selective mechanism is compatible with DETR-like detectors that do not have non-maximum suppression. 
It performs scale attention by comparing the values of the existing features among different scales with SoftMax.
% The motivation of cross-scale pixel attention is to highlight text features in enlarged feature maps. To suppress the irrelevance from the other enlarged areas, it computes the attention weights of multi-scale features across both the scale and spatial dimensions, which implicitly down-weight the useless content. 
More details are illustrated in Fig.~\ref{fig:network}, the feature maps of a square image from the ResNet backbone are with the size $R_1, R_2, R_3, R_4$. We enlarge them to square feature maps with sizes $D_{1}, D_{2}, D_{3}, D_{4}$. Following Deformable-DETR, the four feature maps are transformed to have the same channels by $1\times 1$ convolutions. Then we re-scaled them to have the same size $D$ and assemble them to obtain $\mathbf{F} \in \mathbb{R}^{D \times D \times C \times 4}$. After that, we use a SoftMax layer to compute the attention map by $\mathbf{A}^{ijk} = \mbox{softmax}([\mathbf{F}^{ijk1}, \mathbf{F}^{ijk2}, \mathbf{F}^{ijk3}, \mathbf{F}^{ijk4}]), \quad i\in [1,D ], \quad j \in [1, D], \quad k\in [1, C]$ for each pixel and each channel.
The attention map $\mathbf{A}$ and the feature map $\mathbf{F}$ have the same shape, \emph{i.e.}, $\mathbf{A}, \mathbf{F} \in \mathbb{R}^{D\times D \times C \times 4}$. They multiply together to obtain the enhanced feature map $\mathbf{F}'$. We disassemble $\mathbf{F}'$ to four feature maps. All of them are with the size $D$ and the channel dimension $C$. We re-scale their size to be $D_{1}, D_{2}, D_{3}$ and $D_{4}$. It is noteworthy that the whole cross-pixel attention module is parameter-free, which brings no training burdens during gradient back-propagation.

\noindent \textbf{Lightweight deformable transformer.}
The four feature maps from the cross-scale pixel attention module are flattened to be vectors. We concatenate them to a long vector, then feed the vector to the deformable transformer. To predict the parameters of PB, we reduce the layers of the standard transformer encoder and decoder from 6 to 2, which is sufficient to yield competitive results. In a deformable transformer, the reference points attend a small set of key sampling points nearby for each query, which are important to the deformable attention module. We adopt a coarse-to-fine strategy~\citep{DeformableDETR} to generate the reference points for the two decoder layers. 
In the first decoder, we adopt rough 2-d reference points derived from the positional embedding via a linear projection. 
In the second decoder, we combine the same 2-d reference points with a 2-d vectors transformed from the output of the first decoder. 
In particular, the 2-d vectors encode the relative offsets according to the first decoder’s learned non-local dependencies, which help to generate more reasonable reference points for the second decoder layer. After that, a 3-layer MLP generates the PB predictions over the entire image.

\subsection{Loss Function}
\label{sec:scloss}

We leave the part about how to generate four ground truth point sets for the top, bottom, left and right sides from an original annotated polygon annotation in the appendix.
In this section, we introduce the shape-constrained loss to supervise the whole network. The network outputs $N$ different PB parameters for each image, while their correspondences to ground truth contours are unknown. In this section, we first introduce how to compute the similarity between the predicted PB and ground truth contour by shape-constrained loss, then provide the loss function for the whole image based on optimized correspondences solved by bipartite matching. 

\noindent \textbf{The shape constraints for each curve.} 
We first revisit the curve fitting loss without constraints used in the lane detection~\citep{LSTR, CLGo}. The ground truth points of a top or bottom curve are given by:
\begin{equation}
    \hat{\mathcal{P}}= \{\left(\hat{x}_{i}, \hat{y}_{i}\right)\}_{i=0}^{K}, \quad  \hat{x}_{i} = \hat{x}_{0} + \frac{\hat{x}_{K}-\hat{x}_{0}}{K}i, \label{eq:gt}
\end{equation}
where the points are ordered from one end to the other, and the adjacent points for the top and bottom curves have the equal distance. The conventional fitting loss is:
\begin{equation}
    \mathcal{L}_{w/o}(\hat{\mathcal{P}}) = \sum_{i=0}^{K}\|\hat{y}_{i} - f(\hat{x}_{i}) \|_{1}+\|e_{0} - \hat{x}_{0}\|_{1} + \|e_{1} - \hat{x}_{K}\|_{1}.
\end{equation}
The fitting loss for the left and right curves can be obtained by exchanging the $x$ and $y$ variables. However, such a fitting loss is unsuitable for detecting texts with diverse shapes and different positions. It has two limitations: (1) the predicted curve segment is not aligned with the ground truth fitting points piece-by-piece; (2) the shape and range of the curve are independently optimized. As demonstrated in Fig. \ref{fig:shape}, the conventional loss is not sensitive to the length of the curves, therefore the text detector tends to detect curves with inaccurate lengths. 

We consider to impose the shape constraints. Points on the predicted curve are sampled according to both curve shape and range:
\begin{equation}
\label{eq:sampling}
\mathcal{P} = \{(x_{i},f(x_{i})) \}_{i=0}^{K}, \quad  x_{i}=e_{0}+\frac{e_{1}-e_{0}}{K}i. 
\end{equation}
Then we compare predicted points $\mathcal{P}$ and ground truth points $\hat{\mathcal{P}}$:

\begin{figure*}[t]
\begin{center}
\includegraphics[width=139mm]{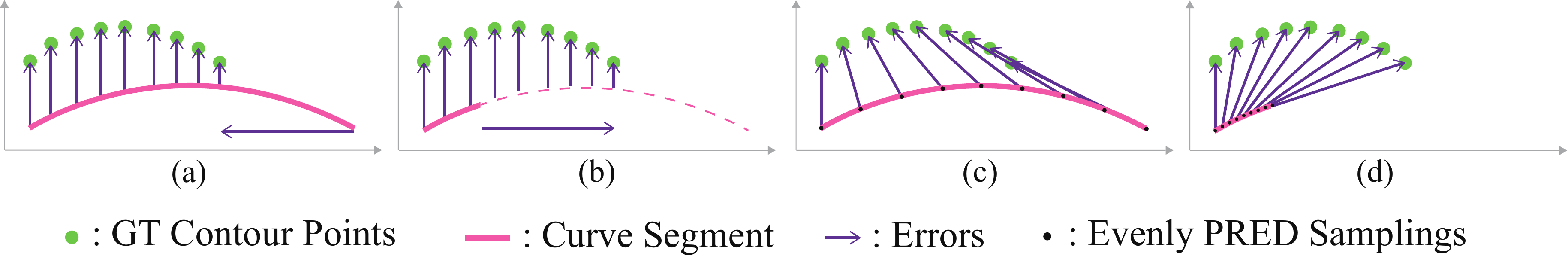}
\vspace{-1.1em}
\end{center}
\caption{\textbf{Diagram of shape constraints.} All curve segments have the same curve coefficients. (a) and (c) have same ranges, so do (b) and (d). Without shape constraints, (a) and (b) show how to compare predicted curve segment with ground truth contour points. (c) and (d) illustrate the way with shape constraints.}
\label{fig:shape}
\vspace{-1.1em}
\end{figure*}

\begin{equation}
\label{eq:pbformer}
    \mathcal{L}(\mathcal{P}, \hat{\mathcal{P}}) = \sum_{i=0}^{K}\|x_{i}-\hat{x}_{i}\|_{1} +\|f(x_{i})-\hat{y}_{i}\|_{1}.
\end{equation}
Leveraging Eq.~\ref{eq:pbformer} in text detection encourages PB to reconstruct the correct length of the contours.

\noindent \textbf{The bipartite matching for the whole image.}
Let the network output of one image be $\mathcal{H} = \{h_{j}\!=\!(c_{j}, \theta_{j}) \}_{j=1}^{N}$, where $c_{j}$ is the confidence score indicating the possibility of a PB covering a text and $N$ is set to be larger than the maximum number of texts in an image.  After sampling the points on the four curves according to Eq.~\ref{eq:sampling}, $\mathcal{H}$ can be further represented by: $\mathcal{H}= \{ h_{j}=(c_{j}, \mathcal{P}_{j}^{t},\mathcal{P}_{j}^{b}, \mathcal{P}_{j}^{l},\mathcal{P}_{j}^{r})\}_{j=1}^{N}$.

For bipartite matching, we pad the ground truth set $\hat{\mathcal{H}}$ with non-text instances to have a size $N$. The element having text instance is represented by $\hat{h}_{j} = (\hat{c}_{j}, \hat{\mathcal{P}}_{j}^{t},\hat{\mathcal{P}}_{j}^{b},\hat{\mathcal{P}}_{j}^{l},\hat{\mathcal{P}}_{j}^{r})$. In particular, $\hat{\mathcal{P}}_{j}^{t},\hat{\mathcal{P}}_{j}^{b},\hat{\mathcal{P}}_{j}^{l},\hat{\mathcal{P}}_{j}^{r}$ are sampled points according to Eq.~\ref{eq:gt}, while they are not need to be instantiated in the matching cost for non-text instances thus are set to be $\emptyset$. $\hat{c}_{j}$ is set to be $1$ for the text and $0$ for the non-text class. We formulate a bipartite matching problem to find an optimal injective function $g:  \hat{\mathcal{H}} \rightarrow \mathcal{H}$, \emph{i.e.}, $g(i)$ is the index of the PB assigned to fitting the $i$-th ground truth text:
\begin{equation}
\label{matchingproblem}
    g^* = \mathop{\arg\min}_{g} \sum_{j=1}^N \mathcal{L}^{fit}\left(\hat{h}_j, h_{g\left(j\right)}\right) + \mathcal{C}^{focal}\left(\hat{c}_j, c_{g\left(j\right)}\right),
\end{equation}
where $\mathcal{L}^{fit}$ is the fitting loss and $\mathcal{C}^{focal}$ is the focal cost. The fitting loss compares the predicted contour and ground truth contour by using the loss defined in Eq.~\ref{eq:pbformer}:
\begin{equation}
\begin{split}
\label{fittingLoss}
& \mathcal{L}^{fit}\left(\hat{h}_j, h_{g\left(j\right)}\right)  = \mathbb{I}_{\hat{c}_j > 0}   \\  
& \left(\mathcal{L}(\hat{\mathcal{P}}_{j}^{t},\mathcal{P}_{g\left(j\right)}^{t}) + \mathcal{L}(\hat{\mathcal{P}}_{j}^{b},\mathcal{P}^{b}_{g\left(j\right)})  
 + \mathcal{L}(\hat{\mathcal{P}}_{j}^{l},\mathcal{P}_{g\left(j\right)}^{l}) + \mathcal{L}(\hat{\mathcal{P}}_{j}^{r},\mathcal{P}_{g\left(j\right)}^{r}) \right).
\end{split} 
\end{equation}
Then, the focal cost is defined as the difference between the positive and negative costs:
\begin{equation}
\begin{split}
\label{focalCost}
& \mathcal{C}^{focal}\left(\hat{c}_j, c_{g\left(j\right)}\right) =  \mathbb{I}_{\hat{c}_j > 0}  \\ 
& \lambda \left[-\alpha \left(1-c_{g\left(j\right)}\right)^{\gamma} \log c_{g\left(j\right)} + \left(1-\alpha\right) c_{g\left(j\right)}^{\gamma} \log \left(1-c_{g\left(j\right)}\right) \right],
\end{split} 
\end{equation}
where $\alpha$ and $\gamma$ are the hyper-parameter for the focal loss. $\alpha$ is used to address the class imbalance, and $\gamma$ adjusts the rate at which easy examples are down-weighted. $\lambda$ adjusts the weight of the focal cost. The bipartite problem (Eq.~\ref{matchingproblem}) can be efficiently solved by the Hungarian algorithm.

\noindent \textbf{Overall Loss.} With the optimized $g^{*}$, the overall loss function is given by:
\begin{equation}
\label{overalloss}
    \mathcal{L}^{overall} =  \sum_{j=1}^N \mathcal{L}^{fit}\left(\hat{h}_j, h_{g^*\left(j\right)}\right) + \mathcal{L}^{focal}\left(\hat{c}_j, c_{g^*\left(j\right)}\right),
\end{equation}
where the $\mathcal{L}^{focal}\left(\hat{c}_j, c_{g^*\left(j\right)}\right)$ is the focal loss:
\begin{equation}
\begin{split}
\label{focalLoss}
     \mathcal{L}^{focal}\left(\hat{c}_j, c_{g^*\left(j\right)}\right) = \lambda [ & - \mathbb{I}_{\hat{c}_j > 0} \alpha \left(1-c_{g^*\left(j\right)}\right)^{\gamma} \log c_{g^*\left(j\right)} \\
    & - \mathbb{I}_{\hat{c}_j =0} \left(1-\alpha\right) c_{g^*\left(j\right)}^{\gamma} \log \left(1-c_{g^*\left(j\right)}\right) ].
\end{split}
\end{equation}
$\alpha$, $\lambda$ and $\gamma$ are the same with the ones in Eq.~\ref{focalCost}.

\section{Experiments}

\definecolor{brilliantlavender}{rgb}{0.96, 0.73, 1.0}
\definecolor{celadon}{rgb}{0.67, 0.88, 0.69}
\definecolor{columbiablue}{rgb}{0.61, 0.87, 1.0}
\begin{table*}[h]
\renewcommand\arraystretch{0.5}
\begin{center}
% \footnotesize
\setlength{\tabcolsep}{3.31mm}{
\caption{Detection results on CTW1500 and Total-Text with or without pre-training on text datasets. Methods with * means training with character-level annotations. TT, MLT, ST, ArT, and CST are abbreviations for Total-Text, MLT2017, SynthText, ArT 2019 and CurvedSynthText which are commonly used pre-training datasets. The blue and green blocks represent the best performance with and without pre-training respectively.}
\label{tab:overallresult}
\begin{tabular}{llccccccccc}
\toprule
\multirow{2}{*}{Method} 
& \multirow{2}{*}{Rep.} & \multirow{2}{*}{PT} & \multicolumn{4}{c}{CTW1500} & \multicolumn{4}{c}{Total-Text} \\ 
& & & F. & Prec. & Rec. & FPS & F. & Prec. & Rec.  & FPS \\
\midrule

TextSnake~\cite{TextSnake} 
& Seg & ST & 75.6 & 67.9 & 85.3 & - & 78.4 & 82.7 & 74.5 & - \\
CRAFT~\cite{CRAFT} 
& Seg  & ST  & 83.5 & 86.0 & 81.1 & - & 83.6 & 87.6 & 79.9 & - \\
SAE~\cite{SAE} 
& Seg & ST  & 80.1 & 82.7 & 77.8 & - & - & - & - & - \\
MSR~\cite{MSR} 
& Seg & ST & 81.5 & 85.0 & 78.3 & 4.3 & 79.0 & 83.8 & 74.8 & 4.3 \\
SAST~\cite{SAST}
& Seg & ST & 81.0 & 85.3 & 77.1 & - & 80.2 & 83.8 & 76.9 & - \\
PSENet~\cite{PSENet}          
& Seg & MLT  & 82.2  & 84.8  & 79.7  & 3.9   & 80.9 & 84.0 & 78.0 & 3.9   \\
PAN~\cite{PAN}             
& Seg & ST   & 83.7  & 86.4  & 81.2  & 39.8  & 85.0 & 89.3 & 81.0 & 39.6     \\
DB~\cite{DBNet}              
& Seg & ST   & 83.4  & 86.9  & 80.2  & 22    & 84.7 & 87.1 & 82.5 & 32       \\
DRRG~\cite{DRRG}            
& Seg & MLT  & 84.5  & 85.9 & 83.0   & -     & 85.7 & 86.5 & 84.9 & - \\
DB++~\cite{DBNetpp}              
& Seg & ST   & 85.3  & 87.9  & 82.8  & 26    & 86.0 & 88.9 & 83.2 & 28       \\
\midrule

LOMO~\cite{LOMO}
& Pts & ST & 78.4 & 89.2 & 69.6 & 4.4 & 81.6 & 88.6 & 75.7 & 4.4 \\
CSE~\cite{CSE}
& Pts & $\times$  & 78.4 & 81.1 & 76.0 & 0.4 & 80.2 & 81.4 & 79.1 & 0.4 \\
Mask-TTD~\cite{MaskTTD}
& Pts & $\times$  & 79.4 & 79.7 & 79.0 & - & 76.7 & 79.1 & 74.5 & - \\
Mask-TextSpotter*~\cite{MaskTextSpotter}
& Pts & ST  & - & - & - & - & 85.2 & 88.3 & 82.4 & - \\
TextRay~\cite{TextRay}         
& Pts & ArT  & 81.6  & 82.8  & 80.4  & 3.2   & 80.6 & 83.5 & 77.9 & 3.5        \\
ContourNet~\cite{ContourNet}      
& Pts & $\times$ & 83.9 & 83.7 & 84.1 & 3.8   & 85.4 & 86.9  & 83.9 & 3.8 \\
% TESTR~\cite{TESTR}           
% & Pts & \small CST+MLT+TT  & 86.6 & 90.8 & 82.8 & 7.3 & 86.2 & 92.4 & 80.7 & 6.9 \\
TESTR*~\cite{TESTR}           
& Pts & \small CST+MLT+TT  & 87.1 & 92.0 & 82.6 & 5.6 & 86.9 & 93.4 & 81.4 & 5.3 \\
TPSNet~\cite{TPSNet}
& Pts & $\times$ & 85.9 & 88.1 & 83.7 & 17.9 & 86.6 & 89.2 & 84.0 & 14.3 \\
TPSNet~\cite{TPSNet}
& Pts & CST  & 86.4 & 87.7 & 85.1 & 17.9 & 88.1 & 89.5 & 86.8 & 14.3 \\
TPSNet*~\cite{TPSNet}
& Pts & CST  & 87.5 & 88.7 & 86.3 & 17.9 & 88.5 & 90.2 & 86.8 & 14.3 \\
\midrule

ATRR~\cite{ATRR}
& Seg+Pts & $\times$  & 80.1 & 80.1 & 80.2 & 10.0 & 78.5 & 80.9 & 76.2 & 10.0 \\
PCR~\cite{PCR}
& Seg+Pts & MLT & 84.7 & 87.2 & 82.3 & 11.8 & 85.2 & 88.5 & 82.0 & - \\
FCENet~\cite{FCENet}          
& Seg+Pts  & $\times$ & 85.1 & 88.1 & 82.3  & 2.7  & 85.8 & 89.3  & 82.5  & 2.9     \\
Boundary*~\cite{BoundaryNet}
& Seg+Pts & ST  & - & - & - & - & 87.0 & 88.9 & 85.0 & - \\
ABCNetV2*~\cite{ABCNetv2}
& Seg+Pts & C+M  & 84.7 & 85.6 & 83.8 & - & 87.0 & 90.2 & 84.1 & - \\
TextBPN~\cite{TextBPN}         
& Seg+Pts  & $\times$ & 84.0 & 87.7 & 80.6 & 12.1  & 86.9 & 90.8  & 83.3 & 10.6 \\
TextBPN~\cite{TextBPN}          
& Seg+Pts & MLT  & 85.0  & 86.5 & 83.6   & 12.2  & 87.9 & 90.7 & 85.2 & 10.7 \\
\midrule

ABCNet~\cite{ABCNet}          
& BezPts & CST  & 81.4  & 84.4  & 78.5  & 6.8   & 84.5 & 87.9 & 81.3 & 6.9 \\
FewBetter~\cite{FewCouldBeBetter}     
& BezPts & CST  & 85.2  & 88.1  & 82.4  & -  & 88.1 & 90.7 & 85.7 & -   \\
% TESTR~\cite{TESTR}           
% & BezPts &  \small CST+MLT+TT  & 85.9  & 90.6  & 81.6  & 7.3 & 87.4  & 92.4  & 82.8 & 6.9    \\
TESTR*~\cite{TESTR}            
& BezPts & \small CST+MLT+TT  & 86.3  & 89.7  & 83.1  & 5.6 & 88.0  & 92.8  & 83.7 & 5.5   \\
\midrule
PBFormer
& PB & $\times$ & \colorbox{celadon}{87.0} & 89.6 & 84.5 & 24.7 & \colorbox{celadon}{87.1}  & 92.1  & 82.6  & 24.6 \\
% PBFormer     
% & PB & CST & \colorbox{columbiablue}{88.0}  & 90.6  & 85.4  & \colorbox{celadon}{24.7}  & \colorbox{celadon}{88.1}  & 93.2  & 83.5 & 24.6  \\
PBFormer     
& PB & CST & \colorbox{columbiablue}{88.7}  & 91.1  & 86.1  & 24.7  & \colorbox{columbiablue}{88.8}  & 93.8  & 84.0 & 24.6  \\
\bottomrule

\end{tabular}}
\end{center}
\end{table*}

\noindent \textbf{Datasets.} 
\textbf{CTW1500}~\citep{CTW1500} is a multi-oriented and curved scene text detection benchmark containing 1,000 training and 500 testing images. Annotations are based on the text-line level with fixed fourteen points. The majority of text instances are curved.
\textbf{Total-Text}~\citep{TotalText} is an another multi-oriented and curved scene text benchmark, while it consists of various text shapes such as multidirectional quadrilateral. It has 1255 training images and 300 testing images. Each instance is annotated by ten point text-line. 
\textbf{DAST1500} \cite{dast1500} has dense and irregular text detection data gathered from online sources. It has 1038 training and 500 testing images, with polygon annotations at the text line level.
\textbf{MLT2019}~\cite{mlt2019} is a multi-oriented multi-language dataset. It features 10 languages from 7 scripts: Chinese, Japanese, Korean, English, French, Arabic, Italian, German, Bangla, and Hindi (Devanagari). The dataset has 10,000 training images, 2,000 validation images, and 10,000 testing images. 
\textbf{ArT2019}~\cite{art2019} is a complex, multi-lingual, arbitrary shape scene text detection dataset. It has 5,603 training and 4,563 testing images with polygon annotations featuring an adaptive number of key points for the text regions.

\noindent \textbf{Evaluation Metrics.} 
We follow the standard metrics F-measures, recall, and precision to evaluate the performance. A prediction is considered as a true positive only when its IoU from the corresponding ground truth contour is larger than 0.5.

\noindent \textbf{Implementation Details.}
The input image size is set to be $800 \times 800$ for training and testing. Loss coefficient $\alpha$, $\gamma$ and $\lambda$ are set as 0.25, 2 and 2. The fixed number of output $N$ is 300. In the cross-scale pixel attention module, $R_1, R_2, R_3, R_4$ are 100, 50, 25, 13, $D_1, D_2, D_3, D_4$ are set as 128, 64, 32, 16, and we set $D=D_2$. For training from scratch, the learning rate is set to be $1 \times 10^{-4} $ and decays by a factor of ten during the last fifth of training. The training process takes about 2 days on 4 Tesla V100 GPUs with the image batch size of 14. For training with pre-training, we pre-train the model for 50 epochs, then fine-tune the model on other datasets' training sets by the same setting as training without pre-training states.
\vspace{-1.2em}

\subsection{Comparison with State-of-the-art Methods}

\begin{figure*}[t]
\includegraphics[width=164mm]{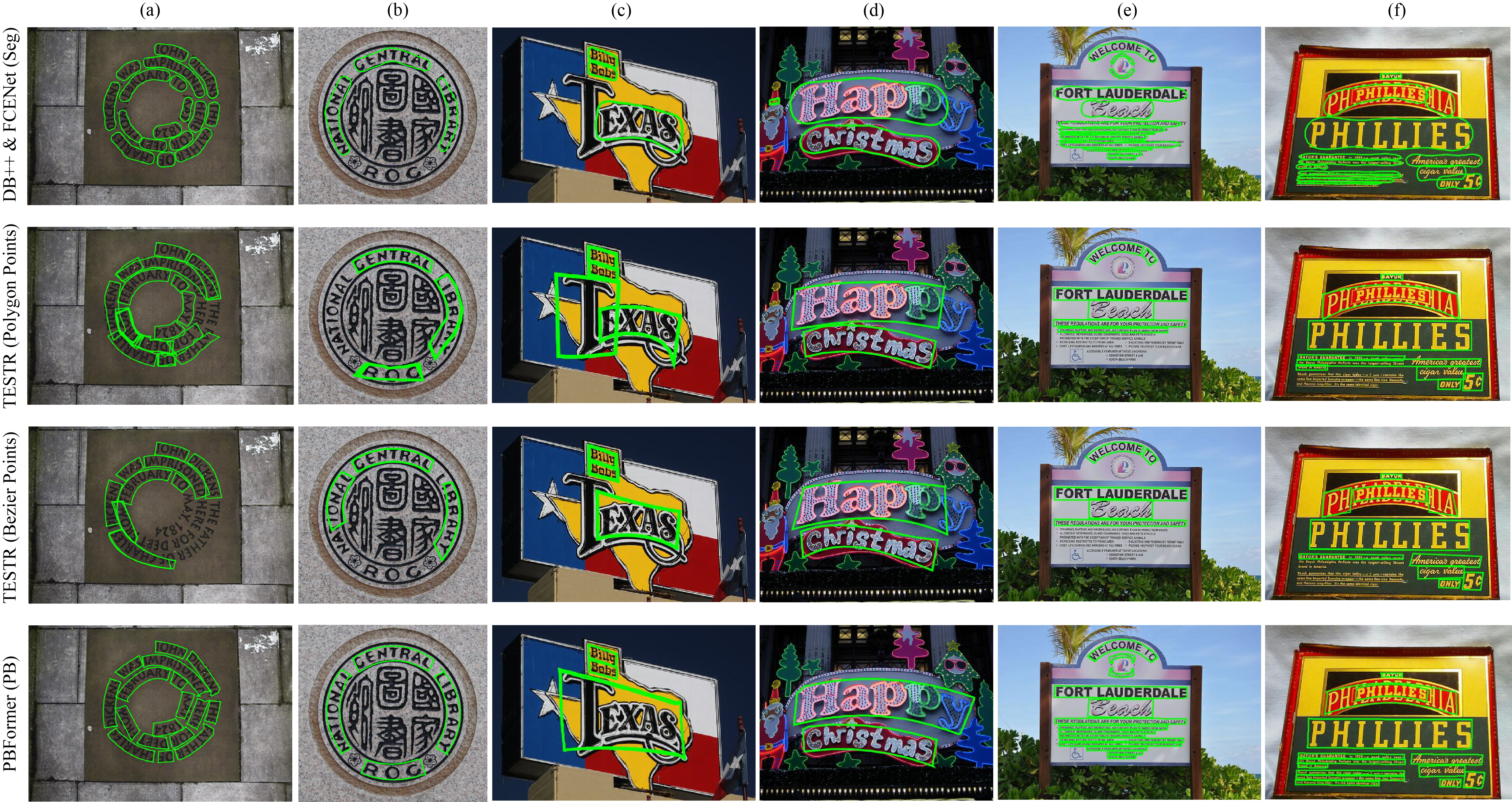}  
\vspace{-1.1em}
\caption{\textbf{Qualitative comparisons with previous SOTA on Total-Text and CTW1500.} Compared to DB++ and FCENet, our PBFormer predicts more compact and precise contours for crowded texts (the first two are DB++'s Total-Text detections, and the last four are FCENet's CTW1500 detections, because they did not release the model of another dataset). Compared to TESTR*, PBFormer reduces false negatives and performs better for long and curved texts.}
\label{fig:viz}
\end{figure*}

\begin{figure*}[t]
\includegraphics[width=164mm]{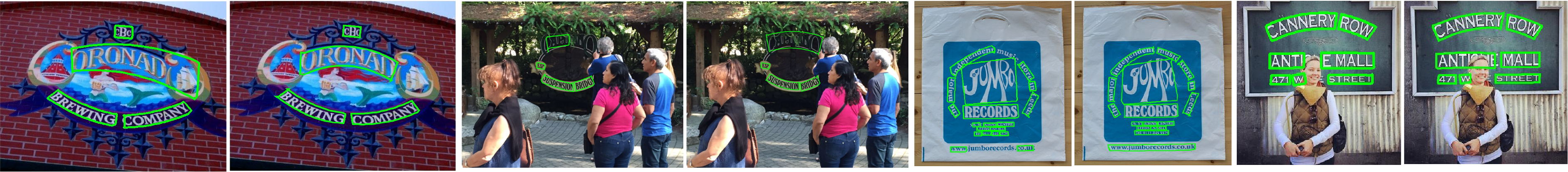} 
\vspace{-1em}
\caption{\textbf{Effect visualization of shape-constrained loss}. For each image pair, the left image shows the results with the fitting loss Eq.~\ref{eq:pbformer}, and the right image is with the shape-constrained loss Eq.~\ref{overalloss}. With the shape-constrained loss, PBFormer outputs more complete contours.}
\label{fig:loss}
\end{figure*}

\begin{figure*}[t]
\includegraphics[width=164mm]{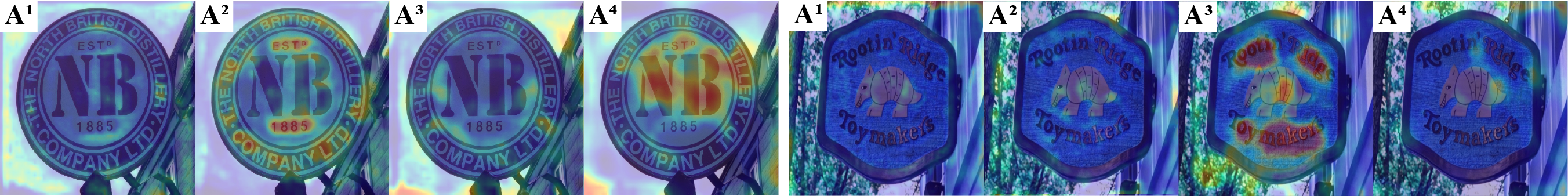} 
\vspace{-1em}
\caption{\textbf{Visualization of CPA's attention maps.}The left four images show CPA produces attention for the small texts at a swallow layer and the large texts at a deep layer. The right four image show attention is concentrated on one layer due to texts having similar sizes.}
\label{fig:CPA}
\end{figure*}

To demostrate the effectiveness of our method, we take methods based on points, segmentation, Bezier points and hybrid representations as competitors. As shown in Tab.~\ref{tab:overallresult}, compared with models trained from scratch, PBFormer establishes a new state-of-the-art of \textbf{87.0}\% on CTW1500, which is \textbf{1.1}\% better than previous best TPSNet while achieving \textbf{1.4} $\times$ FPS. Moreover, PBFormer yields the best F-measure \textbf{87.1}\% on Total-Text, which is \textbf{0.2}\% better than previous best TextBPN while being \textbf{2.3} $\times$ FPS. These prove that PBFormer is easy to train and get the superior detection performance.

As for models with pre-training, previous methods have different choices of pre-training datasets, so we choose to pre-train on single CurvedSynthText as another Transformer-based FewBetter~\cite{FewCouldBeBetter} did. Note that we do not take advantages from character-level annotations to boost the detection performance like TPSNet* and TESTR*.
As shown in Tab.~\ref{tab:overallresult}, compared with models with pre-training, PBFormer can achieve better results from the model pre-training. On the CTW1500 dataset, PBFormer achieves the best results. It outperforms previous best performance by \textbf{1.2}\% in terms of F-measure and is \textbf{1.4} $\times$ faster than TPSNet*. Compared to the Transformer-based FewBetter and TESTR*, PBFormer boost the F-measure by \textbf{3.5}\% and \textbf{1.6}\%, while being faster than TESTR* \textbf{4.4} $\times$.
On the Total-Text dataset, PBFormer performs a \textbf{1.7} $\times$ FPS while being a slight \textbf{0.3}\% higher F-measure compared to TPSNet*. Compared to the Transformer-based FewBetter and TESTR*, PBFormer boost the F-measure by \textbf{0.7}\% and \textbf{0.8}\%, and being faster than TESTR* \textbf{4.5}~$\times$.

We also evaluate our model on DAST1500, which is not only arbitrary-shaped but also dense scene text detection dataset. In Tab.~\ref{tab:dast1500}, our PBFormer improve the F-measure by 0.5\% compared with two-stage Mask-RCNN based method, showing PB representation the whole efficient pipeline without NMS is adaptive for dense scenario. Moreover, in Tab.~\ref{tab:mltart2019}, PBFormer also demonstrates improvement, proving the adaptive ability to large scale datasets.

\begin{table}[t]
\renewcommand\arraystretch{0.5}
\begin{center}
\setlength{\tabcolsep}{4.5mm}{
\caption{Experiments on MLT2019 and ArT2019.}
\label{tab:mltart2019}
\begin{tabular}{lclc}
\toprule
\multicolumn{2}{c}{MLT2019} & \multicolumn{2}{c}{ArT2019} \\
Method         & F. & Method & F.  \\
\midrule
PSENet~\cite{PSENet}   & 65.8  & TextRay~\cite{TextRay} & 66.2 \\
CRAFTS*~\cite{CRAFTS}  & 68.1  & PCR~\cite{PCR}         & 74.0 \\
DBNet++~\cite{DBNetpp} & 71.4  & TPSNet~\cite{TPSNet}   & 78.4 \\
PBFormer & \colorbox{columbiablue}{73.2}  & PBFormer & \colorbox{columbiablue}{79.7} \\
\bottomrule
\vspace{-2em}
\end{tabular}}
\end{center}
\end{table}

\begin{table}[t]
\renewcommand\arraystretch{0.5}
\begin{center}
\setlength{\tabcolsep}{5.5mm}{
\caption{Comparisons on DAST1500's testing set.}
\label{tab:dast1500}
\begin{tabular}{lccc}
\toprule
Method    & F. & Prec. & Rec.  \\
\midrule
TextBoxes~\cite{TextBoxes} & 50.9 & 67.3 & 40.9 \\
RRD~\cite{RRD}             & 53.0 & 67.2 & 43.8 \\
EAST~\cite{EAST}           & 62.0 & 70.0 & 55.7 \\
SegLink~\cite{SegLink}     & 65.3 & 66.0 & 64.7 \\
CTD+TLOC~\cite{CTW1500}    & 66.6 & 73.8 & 60.8 \\
PixelLink~\cite{PixelLink} & 74.7 & 74.5 & 75.0 \\
ICG~\cite{dast1500}        & 79.4 & 79.6 & 79.2 \\
ReLaText\citep{ReLaText}   & 85.8 & 89.0 & 82.9  \\
MAYOR\citep{XugongMAYOR}   & 86.6 & 87.8 & 85.5 \\
PBFormer                   & \colorbox{columbiablue}{87.1} & 90.6 & 83.9 \\
\bottomrule
\vspace{-4.3em}
\end{tabular}}
\end{center}
\end{table}

\begin{figure}[t]
\begin{center}
\includegraphics[width=85mm]{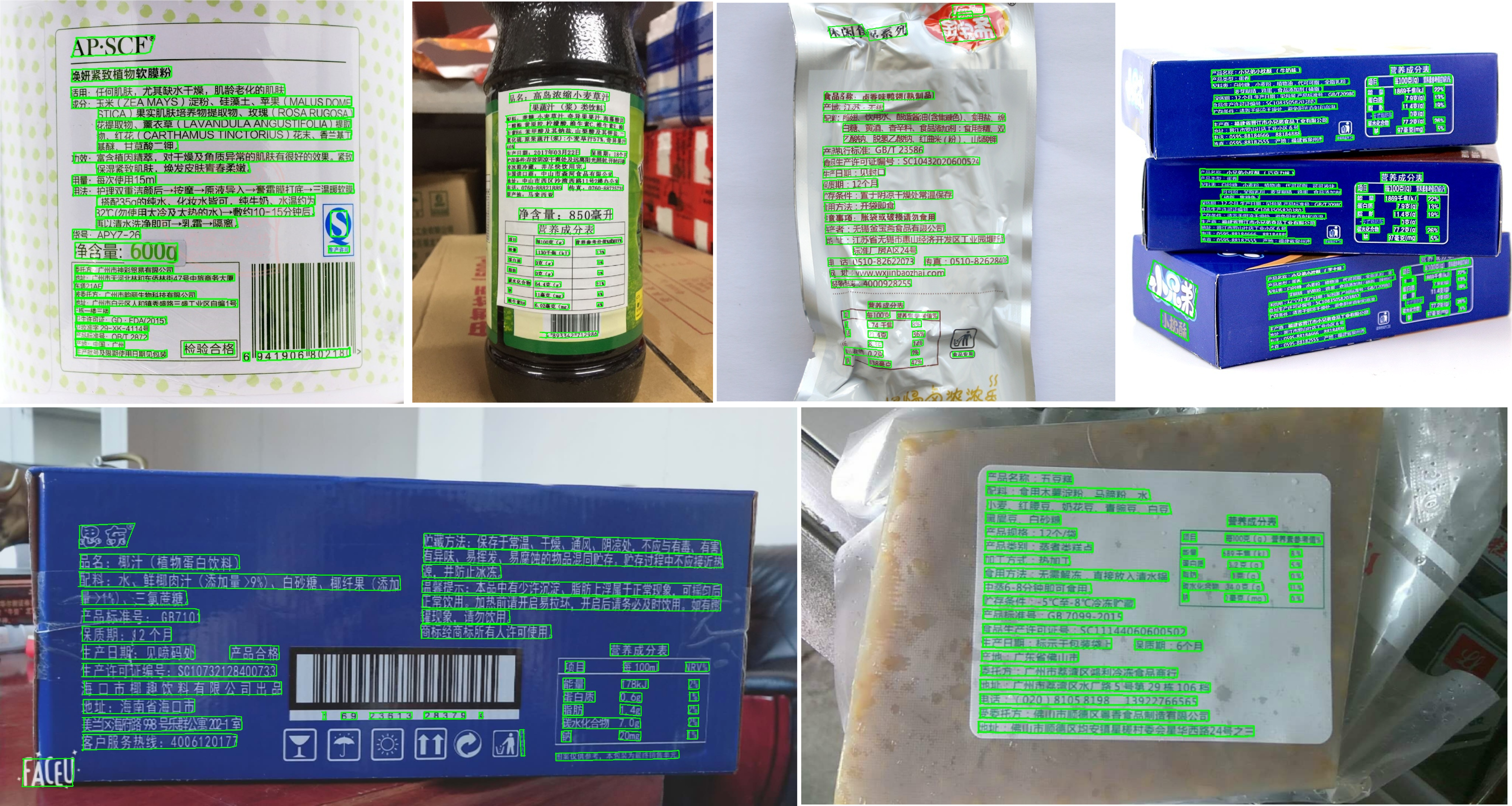}
\end{center}
\vspace{-1.1em}
\caption{\textbf{Qualitative results on the testing set of DAST1500.}}
\label{fig:viz_dast}
\end{figure}

\noindent \textbf{Qualitative results.}
Considering crowded texts in Fig.~\ref{fig:viz}(a),(e), and (f), PBFormer performs fewer false-negatives than TESTR and more accurate contours than DB++ and FCENet. When texts have very long shapes or have characters' large scale-changes, PBFormer detected more completed contours than DB++, FCENet, and TESTR, as Fig.~\ref{fig:viz}(d) and (c) have shown.
Moreover, we visualize detection results on DAST1500's testing test in Fig.~\ref{fig:viz_dast}. As a description of the complete individual of the text, PB is less likely to cause false conglutination. Meanwhile, as an efficient NMS-free pipeline, PBFormer is not easy to mistakenly suppress adjacent detection results.

\begin{table}[t]
\renewcommand\arraystretch{0.5}
\begin{center}
\setlength{\tabcolsep}{5.2mm}{
\caption{Comparison on Total-Text of the same network with different text representations.}
\label{tab:PBFormer}
\begin{tabular}{clccc}
\toprule
CPA & Rep. & F. & Prec. & Rec. \\
\midrule
- & PB  & \textbf{85.8} & \textbf{90.2} & \textbf{81.9} \\
- & Pts & 82.8 & 88.4 & 77.8 \\
- & BezPts & 83.9 & 89.8 & 78.7 \\
\midrule
% - & \checkmark & PB  & \textbf{86.0} & \textbf{90.5} & \textbf{81.9} \\
% - & \checkmark & Pts & 83.2 & 87.6 & 79.1 \\
% - & \checkmark & BezPts & 84.2 & 89.7 & 79.4 \\
% \midrule
% TESTR      
% & two-stage Deformable-DETR & 6 & - & PB  & \textbf{87.4} & -    & - \\
% TESTR         
% & two-stage Deformable-DETR & 6 & - & Pts & 85.3 & 89.7 & 81.2 \\
% TESTR           
% & two-stage Deformable-DETR & 6 & - & Bez & 86.3 & 90.3 & 82.6 \\
% \hline
\checkmark & PB     & \textbf{87.1} & \textbf{92.1} & \textbf{82.6} \\
\checkmark & Pts    & 84.6 & 90.5 & 79.5 \\
\checkmark & BezPts & 85.3 & 90.9 & 80.4 \\
\bottomrule
\vspace{-2em}
\end{tabular}}
\end{center}
\end{table}

\begin{table}[h]
\renewcommand\arraystretch{0.5}
\begin{center}
\setlength{\tabcolsep}{5.2mm}{
\caption{Comparisons with FPN and ASF. Ext. means extended trainable parameters.}
\label{tab:csap}
\begin{tabular}{lcccc}
\toprule
Module     & Ext.       & F.    & Prec. & Rec.  \\
\midrule
CPA        & -          & \textbf{87.1}  & \textbf{92.1}  & \textbf{82.6} \\
ASF        & \checkmark & 85.3 & 90.9 & 80.4 \\
FPN        & \checkmark & 85.0 & 89.5 & 81.0 \\
\bottomrule
% \vspace{-2.7em}
\end{tabular}}
\end{center}
\end{table}

\begin{table}[h]
\renewcommand\arraystretch{0.5}
\begin{center}
\setlength{\tabcolsep}{4.5mm}{
\caption{The impact of CPA's two components.}
% \vspace{-1em}
\label{tab:cpa_study}
\begin{tabular}{lcccc}
\toprule
Enlarge & Attention & F.    & Prec. & Rec.  \\
\midrule
-          & -          & 86.0 & 90.5 & 81.9 \\
\checkmark & -          & 85.2 & 88.9 & 81.8 \\
- & \checkmark          & 86.1 & 90.6 & 82.1 \\
\checkmark & \checkmark & \textbf{87.1}  & \textbf{92.1}  & \textbf{82.6} \\
\bottomrule
% \vspace{-2em}
\end{tabular}}
\end{center}
\end{table}

\begin{table}[t]
\renewcommand\arraystretch{0.5}
\begin{center}
\setlength{\tabcolsep}{1.7mm}{
\caption{Text recognition results when equipping PB predictions with popular recognition modules.}
% \vspace{-1em}
\label{tab:recognition}
\begin{tabular}{clcccc}
\toprule
~ & ~ & \multicolumn{2}{c}{Total-Text} & \multicolumn{2}{c}{CTW1500} \\
Input & Recognition Module                                      & None & Full & None & Full \\
\midrule
$\mathbf{P}_{img}$ & GTC~\cite{GTC}                             & 52.7 & - & 40.3 & - \\
$\mathbf{P}_{img}$ & TROCR~\cite{TROCR}                         & 59.5 & - & 42.9 & - \\
$\mathbf{P}_{feat}$ & MaskTextSpotterV3~\cite{MaskTextSpotterV3}& 68.4 & 76.6 & 50.5 & 77.1 \\
\bottomrule
\vspace{-1em}
\end{tabular}}
\end{center}
\end{table}

\begin{figure}[t]
\begin{center}
\includegraphics[width=85mm]{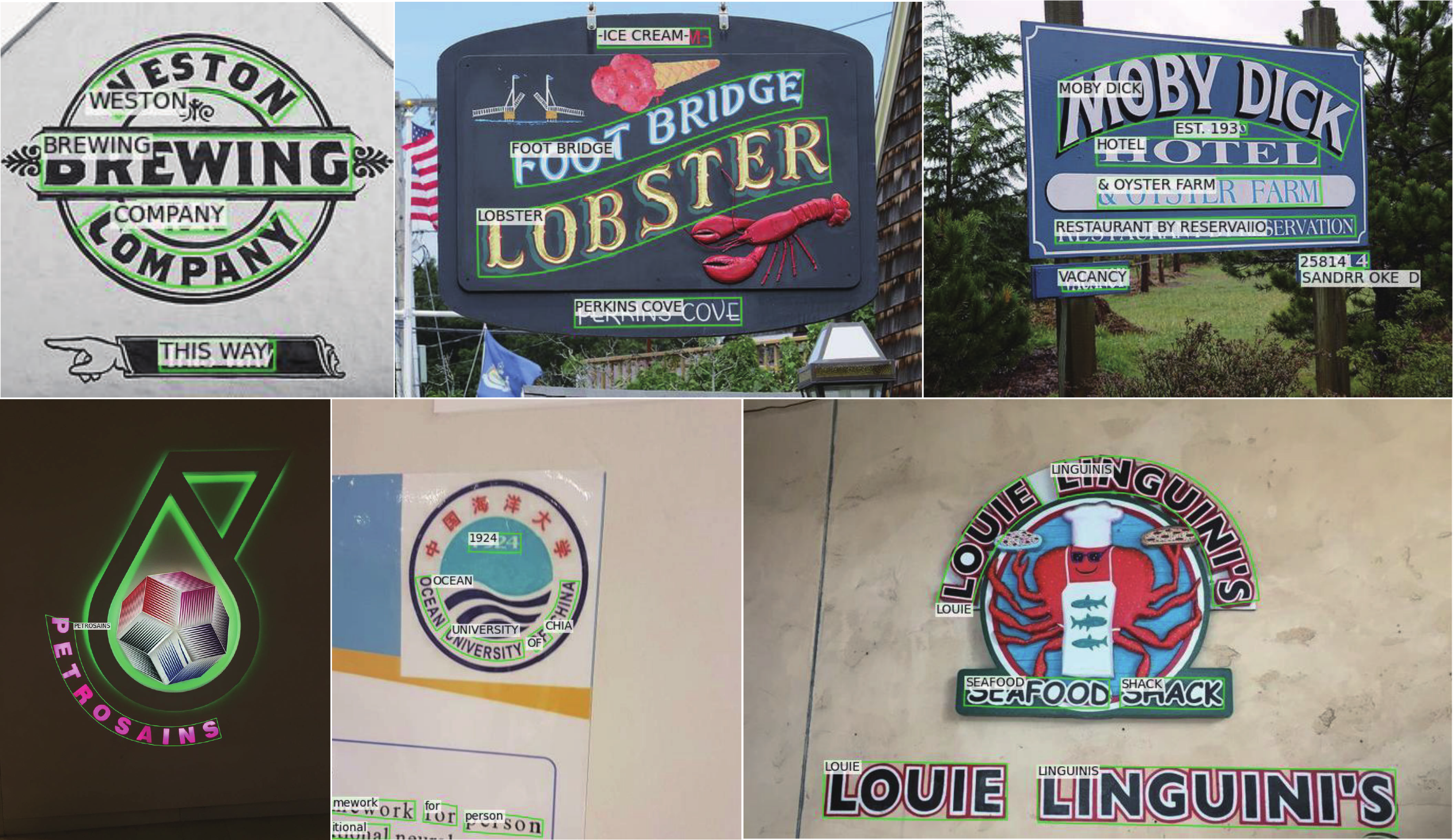}
\end{center}
\vspace{-1.1em}
\caption{\textbf{Visualization of recognition and detection results based on PB on CTW1500 (top) and Total-Text (bottom). We send the detection features represented by PB into the existing recognition module.}}
\vspace{-1.7em}
\label{fig:viz_recognition}
\end{figure}

\subsection{Ablation Study}

\subsubsection{Investigate Polynomial Band Representation}
In this section, we investigate the effects of different text representations under the same network. Tab.~\ref{tab:PBFormer} shows that PB outperforms polygon points and Bezier control points in every network configuration. In the appendix, we also show that PB can distinguish individual texts facing adjacent and overlapping texts better than other representations.

\subsubsection{Investigate Cross-scale Pixel Attention Module}
~

\noindent \textbf{Compare CPA with FPN and ASF.}
To further investigate the effect of the CPA module, we replace CPA with other fusion modules.
In Tab.~\ref{tab:csap}, CPA performs 1.8\% and 2.1\% better than FPN and ASF~\citep{DBNetpp}. We attribute it to (1) FPN will \textbf{NOT} improve the object detection performance because the cross-level feature exchange is already adopted by the multi-scale deformable attention module~\citep{DeformableDETR}. FPN degrades text detection performance because introduced additional parameters are not learned well because the text dataset we used is much smaller than the common object dataset. (2) CPA's selective mechanism is more compatible with DETR-like detectors than ASF. The ASF works better with DB++, which tends to preserve both the global structures and local details for accurate segmentation masks. A detailed comparison can be found at the appendix. 

\noindent \textbf{How CPA enhances multi-scale features?}
CPA contains two operations, enlarging features by a resizing operation and making cross-pixel attention by a softmax layer. In Tab.~\ref{tab:cpa_study}, we can observe (1) just enlarging features performs 0.8\% worse; (2) only using attention has a minor 0.1\% improvement; (3) combining both boosts the performance by 1.1\% significantly. The effectiveness of the CPA module is that the attentional fusion adaptively highlights texts' features at a suitable scale and suppresses the features of other scales. Fig.~\ref{fig:CPA} demonstrates the four attention maps across scales from shallow to deep layers of the backbone.

\noindent \textbf{Applicablity of CPA to different representations.} In order to verify the scalability of CPA to different representations. We investigate the their effects one by one for different representations. In Tab.~\ref{tab:PBFormer}, we observe the improvements of CPA are consistent for different text representations. 

\subsubsection{Effect of Shape-constrained Loss Functions}
Using the shape-constrained loss, the performance on Total-Text increases from 83.5\% to 87.1\%. As Fig.~\ref{fig:loss} shows, the model trained with shape-constrained loss can produce more complete contours (the right image in the pair) than the models without the loss (the left image in the pair).

\subsubsection{Ability to Work with Existing Recognition Networks}
It would be great to validate that a text represented by PB is feasible to current text recognition modules. One way is feeding text image patches $\mathbf{P}_{img}$: we crop a text image patch using the circumscribed bounding box of the PBFormer's detection result, then feed the text image patch into the popular recognition modules GTC~\cite{GTC} and TROCR~\cite{TROCR}; 
The other way is inputting text features $\mathbf{P}_{feat}$: we use Hard RoI masking~\cite{MaskTextSpotterV3} to extract text features inside a text contour generated by sampling PB, then input those features into the recognition module of MaskTextSpotterV3~\cite{MaskTextSpotterV3}. Tab~\ref{tab:recognition} shows the compatibility with other recognition modules no matter extracting images or features. In addition, using MaskTextSpotterV3 achieves a higher performance, because it has been pre-trained on Total-Text and CTW1500, while the other two models have not. Fig.~\ref{fig:viz_recognition} show some qualitative results.

\section{Conclusion}
We have presented PBFormer, an efficient and accurate text detection method. It is superior to handle crowded texts or texts with diverse shapes. PBFormer equips a new text representation, Polynomial Band, to a transformer-based network consisting of a cross-scale pixel attention module and a lightweight deformable transformer. We supervise the network with a shape-constrained loss term, encouraging the network to output the correct contour length. PBFormer shows strong robustness when training without pre-training on the additional datasets, which is much more resource-friendly than other transformer-based methods. 

\bibliographystyle{ACM-Reference-Format}
% \balance
\bibliography{sample-base}

\end{document}